\documentclass{article} 
\usepackage{iclr2027_conference,times}

\usepackage{amsmath,amsfonts,bm}

\def\eqref#1{equation~\ref{#1}}

\def\1{\bm{1}}

\DeclareMathAlphabet{\mathsfit}{\encodingdefault}{\sfdefault}{m}{sl}
\SetMathAlphabet{\mathsfit}{bold}{\encodingdefault}{\sfdefault}{bx}{n}

\usepackage{amsmath}

\iclrfinalcopy

\usepackage[utf8]{inputenc} 
\usepackage[T1]{fontenc}    
\usepackage{hyperref}       
\usepackage{url}            
\usepackage{booktabs}       
\usepackage{amsfonts}       
\usepackage{nicefrac}       
\usepackage{microtype}      
\usepackage{xcolor}         

\usepackage{hyperref}
\usepackage{url}
\usepackage{enumitem}
\usepackage{graphicx}
\usepackage{multirow}
\usepackage{cleveref}
\usepackage{ulem}
\usepackage{multirow}
\usepackage{makecell}
\usepackage{float}
\usepackage{amsthm}
\usepackage{booktabs}
\usepackage{tabularx}
\usepackage{subcaption}
\usepackage{wrapfig}

\title{RoboHarness: Memory-Driven Orchestration of Heterogeneous Robot Policies for Long-Horizon Planning}

\author{\textbf{Jinbang Huang}\textsuperscript{1,$\dagger$},
\textbf{Zhiyuan Li}\textsuperscript{2,$\dagger$,*},
\textbf{Yuanzhao Hu}\textsuperscript{3,*},
\textbf{Ran Qi}\textsuperscript{2,*},
\textbf{Yixin Xiao}\textsuperscript{1},
\textbf{Mark Coates}\textsuperscript{4},\\
\textbf{Tongtong Cao}\textsuperscript{5},
\textbf{Zhanguang Zhang}\textsuperscript{1},
\textbf{Yingxue Zhang}\textsuperscript{1}\\
\textsuperscript{1}Huawei Noah's Ark Lab,
\textsuperscript{2}University of British Columbia,
\textsuperscript{3}University of Toronto,\\
\textsuperscript{4}McGill University,
\textsuperscript{5}Department of Foundation Model, 2012 Labs
}

\usepackage{newunicodechar}
\newunicodechar{，}{,}

\begin{document}

\maketitle

\begingroup
\renewcommand{\thefootnote}{}
\footnotetext{\scriptsize
$^{\dagger}$Equal contribution.
$^{*}$Work done during an internship at Huawei Noah's Ark Lab.
}
\endgroup

\begin{abstract}
Long-horizon robotic tasks require a breadth of capabilities beyond what any single existing robot control policy can reliably provide. Combining heterogeneous policies with complementary strengths offers a promising solution, but introduces two key challenges: uncertain capability boundaries and distribution mismatches during policy handoffs. These challenges remain largely unaddressed by existing planning methods, which typically assume homogeneous, predefined skills with fixed applicability. We propose RoboHarness, a unified framework that encapsulates independently developed heterogeneous policies, including vision-language-action models (VLAs), world-action models (WAMs), reinforcement learning (RL) policies, and task and motion planners (TAMP), as reusable agentic skills. RoboHarness integrates understanding, memory, and evolution skills to reason about policy capabilities and support capability-aware task decomposition and policy routing. To mitigate distribution mismatches during policy handoffs, we introduce Memory Bridge, a plug-in policy-chaining mechanism that enables reliable transitions between heterogeneous policies without joint retraining. Extensive experiments across five public benchmarks, 500 customized tasks across 10 classes, and 135 real-robot trials demonstrate substantial gains in long-horizon and memory-dependent tasks, as well as robustness to out-of-distribution conditions.

\end{abstract}

\section{Introduction}

Long-horizon robotic tasks require instruction alignment, generalization across environmental variations, logical consistency, and noise robustness~\citep{Garrett2020-cr,Yang2024-xo}. Despite substantial progress~\citep{huang2026sct,Black2024-cc,Silver2024-ew,yeunidomain}, no single policy reliably meets all these requirements. Vision-language-action models (VLAs) provide semantic grounding but struggle with long-horizon consistency~\citep{Black2024-cc,kim2025fine,zheng2025x}. World-action models (WAMs) support predictive control but face prediction errors and inference latency~\citep{zhang2026world}. Reinforcement learning (RL) post-training improves closed-loop control but depends on reward design and training coverage~\citep{zang2025rlinf,intelligence2025pi}. Task and motion planning (TAMP) offers logical and geometric reasoning but relies on predefined abstractions and skills~\citep{Silver2023-mi,pmlr-v229-kumar23a,pmlr-v270-liu25d,Huang2025-ue}. These complementary strengths motivate coordinating heterogeneous policies for long-horizon execution.

Such coordination presents two challenges. First, task decomposition and policy selection must account for uncertain, context-dependent capability boundaries. Conventional planners typically assume predefined skills with fixed applicability~\citep{Liang2024-hf,han2024interpret,kaelbling2011hierarchical}, whereas independently developed policies differ in architecture, interfaces, and assumptions~\citep{zhang2026world,garrett2021integrated}. Their applicability may overlap, and relative performance varies with object configurations, observations, instructions, and execution history. Planners must therefore jointly determine task decomposition and policy assignment under current conditions. Second, successful policy selection does not ensure compatible handoffs: one policy's terminal state may fall outside the next policy's feasible execution region or input distribution~\citep{lee2021adversarial,lee2019composing}, causing failures and cascading errors. Effective orchestration thus requires both capability-aware planning and reliable policy handoffs.

We propose RoboHarness, a unified framework that exposes independently trained or hand-designed policies as reusable agentic skills through a common interface accommodating policy-specific inputs and outputs. It uses contextual and historical execution evidence to guide policy assignment, and the incoming policy's retrieved experience to construct compatible handoffs. Our main contributions are:
\textbf{(1)} We introduce RoboHarness, a unified framework that orchestrates heterogeneous robot policies as reusable agentic skills for long-horizon tasks.
\textbf{(2)} We integrate understanding, memory, and self-evolution skills to characterize and adapt context-dependent policy capabilities, supporting capability-aware task decomposition and routing.
\textbf{(3)} We introduce Memory Bridge, a policy-chaining algorithm that learns a local progress ranking from retrieved trajectories and selects reachable, memory-supported handoff configurations without joint policy retraining.
\textbf{(4)} We evaluate RoboHarness across five public benchmarks, 500 customized tasks, and 135 real-robot trials, demonstrating substantial improvements in long-horizon and memory-dependent task success and out-of-distribution robustness.

\section{Related Work}

\textbf{Harness Systems for Agentic and Robotic Execution.}
Harness systems embed language models in execution loops with tools, memory, and feedback for state tracking and iterative refinement~\citep{yao2022react,schick2023toolformer,shinn2023reflexion}. Extensions include executable skill libraries, multi-agent collaboration, and software-engineering agents~\citep{wang2023voyager,wu2023autogen,hong2024metagpt,wang2024executable,yang2024swe,wang2025openhands,xia2024agentless}. Concurrent robotic harnesses support VLA data collection, execution, and self-improvement~\citep{li2026roboclaw,zhang2026harness}, incorporating physical execution constraints~\citep{xiao2026enpire}. Heterogeneous policies additionally differ in capabilities, assumptions, and representations~\citep{zhang2026world,pmlr-v229-dalal23a,huang2026h,pmlr-v205-huang23c}, complicating selection and handoffs. Existing robotic harnesses orchestrate learned primitives~\citep{li2026roboclaw} or combine a VLA with analytic primitives~\citep{zhang2026harness}, but do not explicitly address handoffs among independently developed policy families. RoboHarness combines capability-aware task decomposition and routing with explicit bridging between heterogeneous policies.

\textbf{Long-Horizon Planning and Policy Chaining.}
Long-horizon robot planning includes hierarchical, task and motion (TAMP), skill-based, and language-model-based approaches~\citep{Silver2024-ew,Liang2024-hf,Athalye2024-lo,Garrett2020-cr,pmlr-v270-liu25d,Oswald2024-vz,Huang2025-ue}. TAMP combines symbolic reasoning with geometric feasibility checks over predefined skills~\citep{garrett2021integrated,kaelbling2011hierarchical}, while language-model-based planners generate and adapt plans from natural-language goals~\citep{Zhao2023-wn,Yang2024-xo,NEURIPS2023_f9f54762,chen-etal-2024-prompt,Hu2023-rb}. At the planning level, these approaches typically operate over low-level skills with specified applicability conditions, while compatibility between successive skills is assumed. Such abstractions do not by themselves resolve uncertain capability boundaries or context-dependent routing among independently developed policies. Beyond high-level planning, policy chaining poses a further challenge for hierarchical execution. Skill chaining aligns one policy's terminal states with the next policy's executable region through learned initiation sets~\citep{konidaris2009skill,huang2023value}, transition mechanisms~\citep{lee2019composing,sahni2017learning,mishra2023generative}, or terminal-state regularization~\citep{lee2021adversarial}, typically requiring dedicated training or policy regularization for a given skill set. RoboHarness jointly addresses task decomposition, policy assignment, and handoff adaptation across independently developed policy families through capability-aware planning and memory-guided bridging without joint retraining.

\section{Problem Definition}

We consider long-horizon planning with heterogeneous robot policies \(\Pi=\{\pi_1,\ldots,\pi_N\}\), without task-specific pretraining of the orchestration framework or joint retraining of constituent policies, while allowing online adaptation and experience accumulation during evaluation. Let \(\mathcal{X}\), \(\mathcal{O}\), and \(\mathcal{G}\) denote the system state, global observation, and subtask instruction spaces, respectively. Each policy \(\pi_i\) has an achievable subtask space \(\mathcal{G}_i\subseteq\mathcal{G}\) and a context-dependent applicability region \(\mathcal{A}_i\subseteq\mathcal{O}\times\mathcal{G}_i\), containing observation--subtask pairs under which it can reliably execute. These regions are not explicitly known; policy metadata and execution evidence inform applicability judgments rather than provide an exact representation of \(\mathcal{A}_i\). RoboHarness wraps each policy as an executable skill and maintains policy-specific episodic memory banks \(\mathcal{M}=\{\mathcal{M}_1,\ldots,\mathcal{M}_N\}\), where \(\mathcal{M}_i\) stores successful trajectories of \(\pi_i\). It additionally maintains an execution log \(\mathcal{H}\) that records the complete agent-level interaction trace for both successful and failed executions.

Given a high-level instruction \(I\), an initial state \(\mathbf{x}_0\in\mathcal{X}\), the policy library \(\Pi\), episodic memory \(\mathcal{M}\), and execution log \(\mathcal{H}\), RoboHarness seeks subtasks \(\tau=(g_1,\ldots,g_T)\), policy assignments \(\rho=(\pi_{k_1},\ldots,\pi_{k_T})\), and bridge trajectories \(B=(b_1,\ldots,b_{T-1})\) whose joint execution completes \(I\). Here, \(g_t\in\mathcal{G}\), \(k_t\in\{1,\ldots,N\}\), and \(t\) indexes subtask stages. Let \(o_t\in\mathcal{O}\) denote the observation at the start of stage \(t\), from which the selected policy's native inputs are constructed. A reliable assignment requires \((o_t,g_t)\in\mathcal{A}_{k_t}\). Let \(\bar{o}_t\) denote the terminal observation of \(\pi_{k_t}\). When needed, \(b_t\) starts from the corresponding post-execution state and produces \(o_{t+1}\), aiming to establish \((o_{t+1},g_{t+1})\in\mathcal{A}_{k_{t+1}}\). The problem thus jointly addresses task decomposition, capability-aware policy assignment, and inter-policy bridging to complete \(I\) with achievable subtasks and compatible execution conditions.

\section{Methodology}
\label{sec:method}

\begin{figure}[t]
\centering
\includegraphics[width=\linewidth]{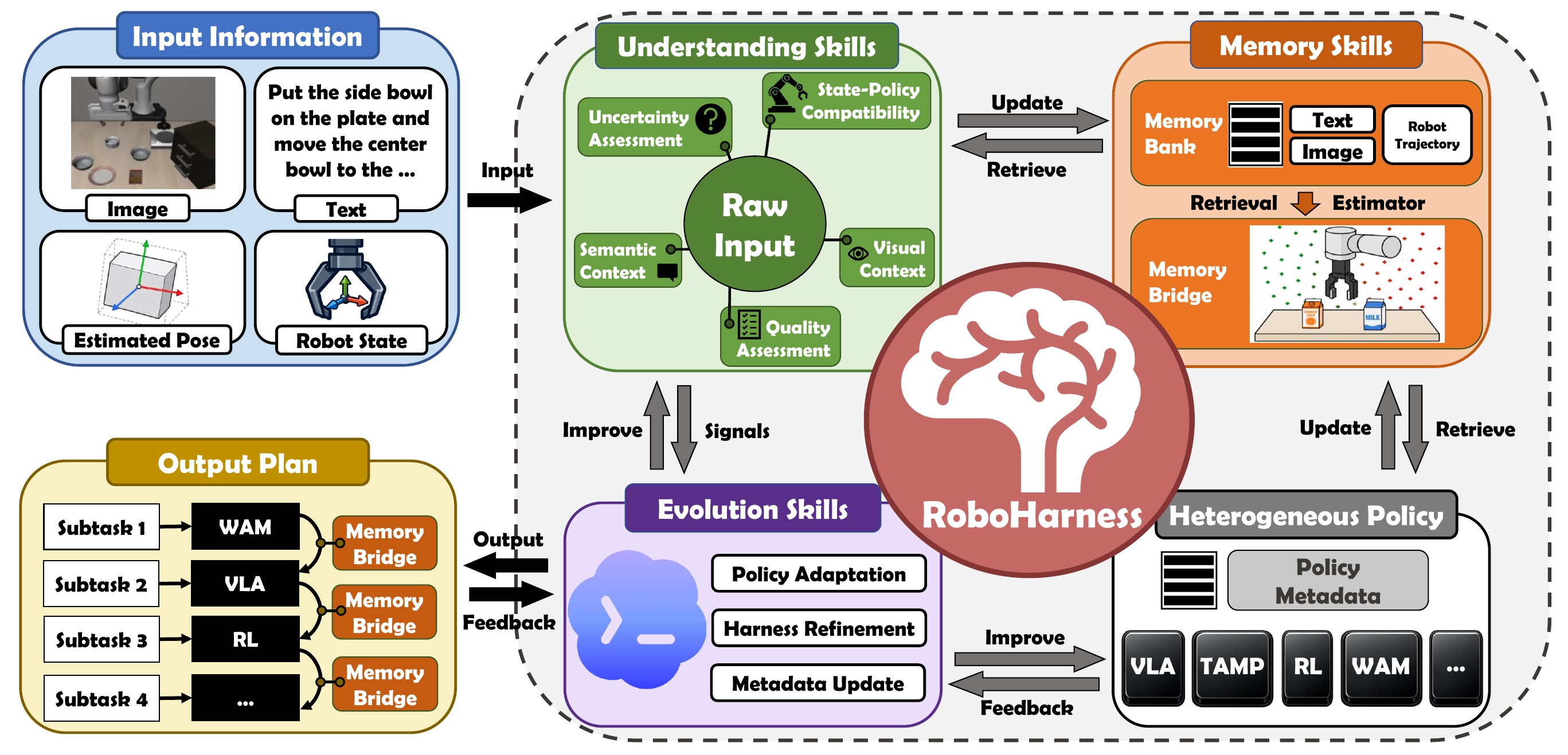}
\vspace{-4mm}
\caption{Overall Framework. RoboHarness integrates heterogeneous robot policies with three auxiliary skill libraries: Understanding Skills, Memory Skills, and Evolution Skills. Understanding Skills transform multimodal observations into task-relevant information for planning and policy selection. Memory Skills retrieve execution evidence for routing and construct Memory Bridge for policy transitions. Evolution Skills use online execution evidence to update individual policies and the harness, forming a closed information loop for capability-aware planning and adaptive orchestration.
}
\vspace{-4mm}
\label{fig:framework}
\end{figure}

RoboHarness uses a coding agent to decompose long-horizon tasks and orchestrate heterogeneous robot policies (\Cref{fig:framework}). Three auxiliary skill libraries support orchestration: understanding guides policy selection; memory retrieves experience and constructs Memory Bridge for policy handoffs; evolution refines policies and orchestration using execution feedback. Policies retain their native interfaces, requiring neither joint retraining nor a shared action representation. Implementation and policy details appear in \Cref{implement,app:underlying_policies}.

\subsection{Understanding Skills}
\label{sec:understanding_skill}

Understanding skills extract decision-relevant information from raw inputs to assess policy applicability in the current context. These inputs include task instructions, visual observations, estimated object poses, robot states, and intermediate execution results. RoboHarness currently includes five types of understanding skills and supports extensions as needed.
\uline{Uncertainty assessment} computes the mean and variance of estimated object poses within a sliding time window to measure temporal stability.
\uline{Visual-context assessment} compares visual embeddings of the current observation with those stored in policy-specific memory.
\uline{Semantic-context assessment} compares text embeddings of task instructions and candidate subtasks with those stored in policy-specific memory.
\uline{Input-quality assessment} evaluates exposure, sharpness, contrast, and image entropy to assess observation quality.
\uline{State-policy compatibility assessment} uses the memory-supported region and local progress estimator constructed by Memory Bridge (\Cref{sec:memory_bridge}) to assess robot-configuration compatibility.
The coding agent invokes these skills on demand and interprets their structured outputs alongside policy cards and the execution log to propose and revise task decompositions and policy assignments. The outputs provide complementary evidence about applicability, rather than a calibrated estimate of policy success. Encoders and assessment procedures are detailed in \Cref{app:understanding_skills}.

\subsection{Memory Skills and Memory Bridge}
\label{sec:memory_skill}

Memory skills serve two roles. First, policy-specific episodic memory supports history-dependent decisions and supplies successful execution experience for Memory Bridge construction. Second, the execution log maintains a complete chronological record of agent and skill interactions and execution feedback across both successful and failed executions. These records provide evidence for capability assessment, failure diagnosis, and evolution.

\subsubsection{Memory Management}
\label{sec:memory_management}

Each policy-specific episodic memory bank \(\mathcal{M}_i\) stores successful trajectories \(\xi=(n_1,\ldots,n_L)\) of temporally linked nodes. Each node \(n\) records a subtask instruction \(g(n)\in\mathcal{G}\), observation \(o(n)\in\mathcal{O}\), robot state \(\mathbf{s}(n)\in\mathbb{R}^{d_s}\), gripper state \(z(n)\), and text and visual embeddings from frozen encoders. The \(d_s\)-dimensional robot state comprises the end-effector pose and joint configuration, extracted from the full system state via \(\mathbf{s}=r(\mathbf{x})\).

Given a subtask \(g\) and observation \(o\), RoboHarness retrieves nodes hierarchically, first by text similarity and then by visual similarity:
\[
\widetilde{\mathcal{N}}
=
\underset{\xi\in\mathcal{M}_i,\;n\in\xi}
{\operatorname{TopK}_{K_{\mathrm{text}}}}
\operatorname{sim}_{\mathrm{text}}\!\left(g,g(n)\right),
\qquad
\mathcal{N}
=
\underset{n\in\widetilde{\mathcal{N}}}
{\operatorname{TopK}_{K_{\mathrm{vis}}}}
\operatorname{sim}_{\mathrm{vis}}\!\left(o,o(n)\right),
\]
where \(\operatorname{TopK}_K\) selects the \(K\) highest-scoring nodes, and both similarity functions use cosine similarity between encoded representations (\Cref{app:understanding_skills}). We denote this operation by \(\operatorname{Retrieve}_{\mathcal{M}_i}(g,o)=\mathcal{N}\). Retrieval may span multiple banks or be restricted to one, such as \(\mathcal{M}_{k_{t+1}}\). During execution, the task and scene context, agent and skill interactions, and execution feedback from both successful and failed executions are recorded chronologically in \(\mathcal{H}\). These complete traces provide evidence for capability assessment, failure diagnosis, and subsequent evolution. Successful policy executions additionally update the corresponding episodic memory bank \(\mathcal{M}_i\) with linked-node trajectories, whereas Memory Bridge retrieves only successful trajectories from \(\mathcal{M}\). Node contents and update procedures are detailed in \Cref{app:memory_data_structure}.

\subsubsection{Memory Bridge}
\label{sec:memory_bridge}

\begin{figure}[t]
\centering
\includegraphics[width=0.95\linewidth]{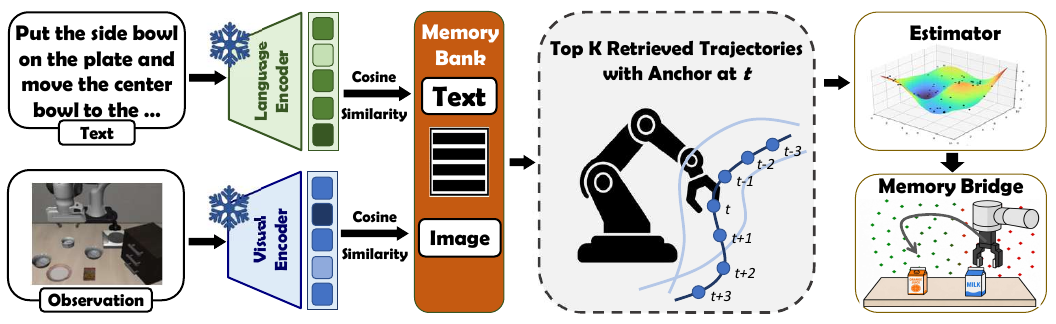}
\vspace{-2mm}
\caption{Memory Bridge. (1) Retrieve top-\(K\) anchors from the next policy's memory using the current observation and upcoming subtask instruction.
(2) Fit a local progress estimator and construct a memory-supported neighborhood from robot states around these anchors.
(3) Generate a bridge trajectory by balancing predicted progress score against motion cost.}
\label{fig:memory_bridge}
\vspace{-4mm}
\end{figure}

Memory Bridge handles transitions where the terminal robot configuration of \(\pi_{k_t}\) is poorly aligned with states supported by \(\pi_{k_{t+1}}\). Its role is to select a handoff configuration using the incoming policy's experience; a motion planner then determines how to reach it. Given the upcoming subtask \(g_{t+1}\) and terminal observation \(\bar{o}_t\), RoboHarness retrieves anchors
\(
\mathcal{N}_t
=
\operatorname{Retrieve}_{\mathcal{M}_{k_{t+1}}}
(g_{t+1},\bar{o}_t)
\)
and generates a bridge trajectory \(b_t\) through three stages.

\textbf{Local progress estimation.}
For each anchor in \(\mathcal{N}_t=\{n_{1,0},\ldots,n_{K,0}\}\), RoboHarness collects neighboring nodes \(n_{i,j}\) by expanding its trajectory bidirectionally while preserving the same subtask. This produces training pairs
\(
(\mathbf{s}_{i,j},y_{i,j})
=
\bigl(\mathbf{s}(n_{i,j}),\Delta t_{i,j}\bigr),
\)
where \(\Delta t_{i,j}\) is the signed temporal offset from the anchor: zero at the anchor, positive for subsequent nodes, and negative for preceding nodes. We train a lightweight progress estimator \(f_{\mathrm{score},t}\) to favor local forward progress, with details of its pairwise ranking SVM implementation provided in \Cref{app:progress_estimator}.

\textbf{Spatial state scoring.}
Candidate handoff states are scored by \(f_{\mathrm{score},t}\) within the supported region
\[
\mathcal{R}_{\mathrm{conf},t}
=
\left\{
\mathbf{s}\in\mathbb{R}^{d_s}
\;\middle|\;
\min_{\bar{\mathbf{s}}\in\mathcal{S}_{\mathrm{ret},t}}
\|\mathbf{s}-\bar{\mathbf{s}}\|_2
\leq\epsilon
\right\},
\]
where \(\mathcal{S}_{\mathrm{ret},t}=\{\mathbf{s}_{i,j}\}\) contains the collected robot states and \(\epsilon>0\) bounds the distance to the nearest sample. We constrain handoff targets to this locally supported region to limit unsupported score extrapolation and guide transitions to the next policy. 

\textbf{Bridge trajectory generation.}
Let \(\mathbf{s}_t\) be the robot state after executing \(\pi_{k_t}\), and \(\mathcal{S}_{\mathrm{plan},t}\) the states reachable from it through feasible motion plans. RoboHarness selects a target by balancing predicted progress against motion cost:
\[
\begin{aligned}
\mathbf{s}^{*}_t
={}&
\arg\max_{\mathbf{s}}
\left[
f_{\mathrm{score},t}(\mathbf{s})
-
\lambda_{\mathrm{motion}}
C_{\mathrm{motion}}(\mathbf{s}_t,\mathbf{s})
\right],\\
\text{subject to}\quad&
\mathbf{s}\in
\mathcal{R}_{\mathrm{conf},t}
\cap
\mathcal{S}_{\mathrm{plan},t},
\quad
f_{\mathrm{score},t}(\mathbf{s})
\geq\theta_{\mathrm{score},t}.
\end{aligned}
\]
where \(C_{\mathrm{motion}}\) measures the control steps required to reach a candidate, \(\lambda_{\mathrm{motion}}\geq0\) weights this cost, and \(\theta_{\mathrm{score},t}\) is the minimum acceptable score. In practice, candidates are sampled, filtered by these constraints, and ranked by the objective.
An off-the-shelf motion controller generates \(b_t\) from \(\mathbf{s}_t\) to \(\mathbf{s}^{*}_t\). Because robot-configuration support alone does not establish task and scene compatibility, the coding agent reassesses the observation and task context after execution before invoking the next policy or revising the routing decision. If the subsequent policy execution succeeds, its linked-node trajectory is additionally written to the incoming policy's episodic memory bank. We initialize \(\theta_{\mathrm{score},t}\) to a preset value and iteratively lower it when no candidate is found, as detailed in \Cref{app:progress_estimator}.

\subsection{Heterogeneous Policies as Agentic Skills}
\label{sec:policy_skill}

RoboHarness wraps each robot policy as a callable agentic skill comprising the policy and an associated policy card. Each policy retains its native implementation, inputs, and outputs without requiring a shared action representation. The policy card records its capabilities, interface requirements, constraints, assumptions, training tasks, and execution statistics. The coding agent uses this metadata and can inspect the implementation code to assess policy applicability and compatibility. As execution evidence accumulates, policy cards are updated to support subsequent planning and assignment. This interface accommodates independently developed policies and supports extension to additional policy families. Our current implementation includes VLAs, WAMs, RL policies, and a TAMP planner. Implementation details and available checkpoint identifiers are provided in \Cref{app:underlying_policies}.

\subsection{Evolution Skills}
\label{sec:evolution_skill}

Evolution skills read evidence from the execution log \(\mathcal{H}\) to update policies, orchestration, and capability descriptions through three mechanisms.
\textbf{Policy Adaptation} adjusts the task models and execution parameters of underlying policies. Our TAMP implementation uses SIMPACT~\citep{liu2026simpact} to refine grasp poses and PDDLLM~\citep{Huang2025-ue} to learn new logical representations from observed interactions.
\textbf{Harness Refinement} uses the coding agent to revise policy interfaces, routing, and inter-skill coordination based on execution traces and runtime errors. It also adjusts orchestration parameters, including retrieval counts and Memory Bridge thresholds, to improve policy selection and handoffs.
\textbf{Metadata Update} revises policy cards using observed successes and failures, recording the objects, tasks, and execution conditions associated with reliable or unsuccessful execution. These updated capability descriptions inform subsequent task decomposition and policy assignment.
Persistent execution failures prompt the coding agent to retrieve relevant log traces and select appropriate updates. The resulting adaptation actions, validation outcomes, and retained changes are written back to \(\mathcal{H}\). Validated changes persist across evaluation episodes, allowing accumulated experience to guide subsequent execution without task-specific pretraining of the orchestration framework. Implementation details, including parameter-tuning procedures, are provided in \Cref{app:evolution_skills}.

\begin{table}[t]
\centering
\caption{LIBERO task following and LIBERO-Plus perturbation robustness. The \emph{Original} column reports performance on the original LIBERO benchmark. Each entry reports the success rate (\%). The \emph{Overall} column aggregates the seven perturbation categories, weighted by their official evaluation instance counts. The best and second-best results are shown in \textbf{bold} and \underline{underlined}, respectively.}
\label{tab:libero_plus}

\small
\setlength{\tabcolsep}{3pt}
\renewcommand{\arraystretch}{1.0}

\resizebox{\linewidth}{!}{%
\begin{tabular}{@{}lc|cccccccc@{}}
\toprule
Method & Original & Robot State & Language & Layout
& Background & Sensor & Camera & Light & Overall \\
\midrule

\(\pi_{0.5}\)
& 96.9 & 77.5 & 85.6 & \underline{85.7}
& 94.6 & 89.7 & 75.4 & \underline{96.9}
& \underline{85.7} \\

\(\pi_0\)
& 91.3 & 40.8 & 63.5 & 76.4
& 84.1 & 80.1 & 61.0 & 89.3
& 69.4 \\

UniVLA
& 95.2 & 46.2 & 69.6 & 31.9
& 81.0 & 21.2 & 1.8 & 69.0
& 42.9 \\

OpenVLA-OFT\(_m\)
& 97.6 & 21.7 & 81.0 & 68.7
& 91.0 & 78.6 & 55.6 & 92.7
& 67.9 \\

X-VLA
& 98.1 & \underline{89.7} & 75.7 & 71.8
& \underline{96.0} & 62.7 & 23.4 & 88.2
& 71.4 \\

Cosmos-Policy
& \underline{98.5} & 63.3 & 81.7 & 82.2
& 88.9 & \textbf{92.7} & \underline{75.8} & 96.5
& 82.2 \\

FAST-WAM
& 97.6 & 44.5 & 68.9 & 60.7
& 53.7 & 37.7 & 16.4 & 78.2
& 51.5 \\

LingBot-VA
& \underline{98.5} & 83.0 & \underline{86.4} & 76.2
& 53.1 & 64.4 & 40.9 & 82.3
& 69.5 \\

\midrule
\textbf{RoboHarness (ours)}
& \textbf{98.7} & \textbf{90.4} & \textbf{97.0}
& \textbf{86.8} & \textbf{97.1} & \underline{90.3}
& \textbf{87.6} & \textbf{97.4} & \textbf{91.9} \\

\bottomrule
\end{tabular}%
}
\vspace{-4mm}
\end{table}

\section{Experimental Setup and Baselines}
\label{setup}

We evaluate RoboHarness against diverse baselines on five public benchmarks and in 135 real-robot trials. The coding agent uses Codex powered by GPT-5.5. Policy checkpoints and skill implementations are detailed in \Cref{implement}. 



\textbf{Heterogeneous Policies Implementations.}
RoboHarness incorporates four independently developed policies: (1) the VLA \(\pi_{0.5}\)~\citep{black2025pi_}; (2) Faster-WAM~\citep{ma2026faster}; (3) an open-source OpenVLA-OFT checkpoint~\citep{kim2025fine}, post-trained with Group Relative Policy Optimization (GRPO)~\citep{shao2024deepseekmath} using RLinf~\citep{zang2025rlinf}; and (4) a TAMP planner designed primarily for geometrically constrained pick-and-place tasks involving predefined objects, with support for multiple motion planners. Implementation details are provided in \Cref{app:underlying_policies}. We inject \(5\text{--}10\%\) noise into simulator-provided object poses to approximate real-world estimation errors. Learned policies use either publicly released checkpoints or checkpoints trained or fine-tuned by us for the corresponding benchmarks. Policy inclusion depends on the availability of compatible checkpoints and the requirements of each benchmark. Each policy's episodic memory bank is initialized with successful trajectories from its corresponding dataset. Benchmark-specific policy configurations are detailed in \Cref{app:policy_benchmark}.

\textbf{Evaluation on Benchmarks.}
The five public benchmarks probe task following, routing behavior and robustness, extended execution horizons, dependence on past observations, and mobile manipulation.
\textbf{LIBERO} measures task-following performance~\citep{liu2023libero}.
\textbf{LIBERO-Plus} assesses out-of-distribution robustness and routing behavior under perturbations~\citep{fei2025libero}.
\textbf{LIBERO-LoHo} assesses extended-horizon execution on tasks averaging approximately four times the original LIBERO horizon~\citep{huang2026h}.
\textbf{RMBench} evaluates memory-dependent long-horizon manipulation at different memory complexity levels~\citep{rmbench}. \textbf{BEHAVIOR-1K} evaluates long-horizon mobile manipulation, for which we conduct a partial evaluation on 6 selected tasks~\citep{li2024behavior1k}. Selected tasks are detailed in \Cref{behavior1k}.
Existing benchmarks do not explicitly require heterogeneous policy switching or evaluate policy handoffs. We introduce \textbf{LIBERO-Complex}, a LIBERO-based benchmark with 500 randomized instances across 10 long-horizon task classes requiring heterogeneous policy coordination, for ablation study (\Cref{task_detail}). During evaluation, execution logs are continuously updated, successful executions update episodic memory banks, and experience is only retained across trials of the same task.

\textbf{Real Robot Experiments.}
We conduct 135 real-robot trials across five assembly task classes. Each task involves 20--30 subtasks, with a 10-minute execution limit per trial. Given language instructions, RoboHarness combines TAMP-based assembly planning with \(\pi_{0.5}\) for contact-rich cabinet opening and exploration. Tasks include constructing short and tall bridges, a Chinese character, and a boat, with some required blocks hidden inside the cabinet to necessitate exploration and retrieval. We test four execution-time disturbances: re-hiding required blocks, partially dismantling structures, perturbing estimated object poses, and introducing distractors.



\begin{figure}[t]
\vspace{-2mm}
\centering
\includegraphics[width=\linewidth]{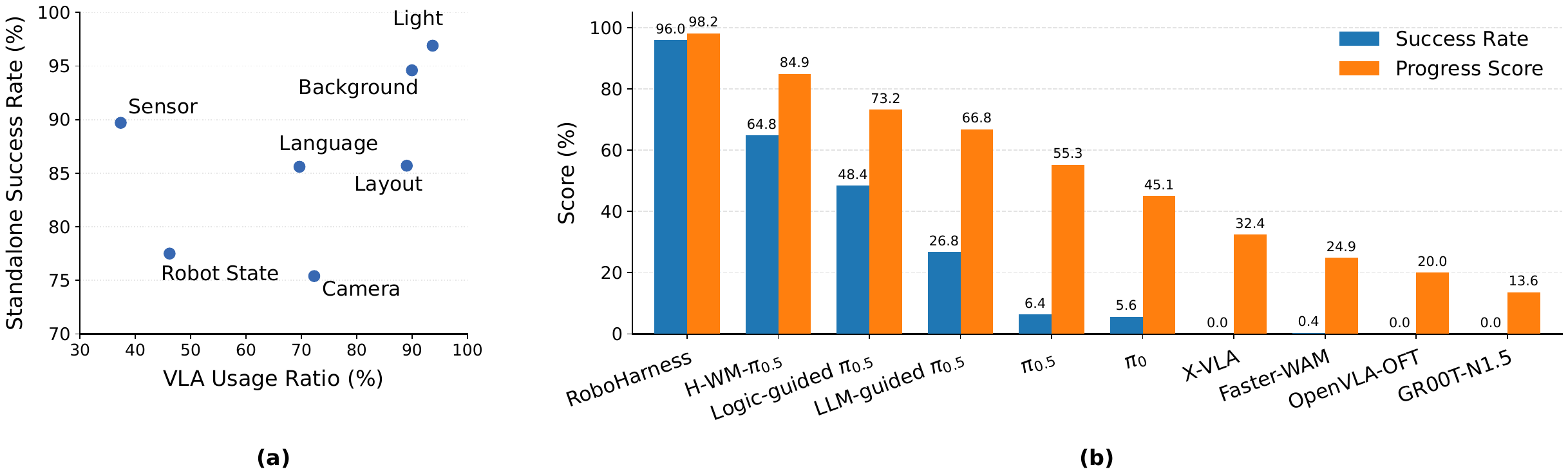}
\caption{(a) Relationship between the VLA invocation ratio and its standalone performance across perturbation categories. (b) LIBERO-LoHo long-horizon evaluation results.}
\label{fig:policy_usage}
\vspace{-4mm}
\end{figure}

\textbf{Baselines} We compare RoboHarness with two families of baselines: \uline{end-to-end policies} and \uline{hierarchical methods}. End-to-end policies include six VLAs, $\pi_{0.5}$~\citep{black2025pi_}, $\pi_0$~\citep{Black2024-cc}, UniVLA~\citep{bu2025univla}, X-VLA~\citep{zheng2025x}, OpenVLA-OFT~\citep{kim2025fine}, and GR00T-N1.5~\citep{nvidia2025gr00tn15}; four world-action models, LingBot-VA~\citep{li2026causal}, Fast-WAM~\citep{yuan2026fast}, Faster-WAM~\citep{ma2026faster}, and Cosmos Policy~\citep{kim2026cosmos}; and Mem-0~\citep{rmbench}, RLC~\citep{larchenko2025behavior}, and G0.5~\citep{liu2026g05}. Hierarchical methods include H-WM~\citep{huang2026h}, LLM-guided VLA~\citep{shi2025}, and logic-guided VLA~\citep{Huang2025-ue}, which provide intermediate guidance for long-horizon execution. Baseline results are taken from published evaluations~\citep{huang2026h,zhang2026world,rmbench,liu2026g05} or obtained by evaluating publicly available checkpoints.


\section{Results}

In this section, we present evaluation results addressing the following six research questions:
\textbf{RQ1:} Does policy usage adapt to context-dependent changes in execution reliability?
\textbf{RQ2:} Can RoboHarness improve robustness to out-of-distribution perturbations?
\textbf{RQ3:} Can RoboHarness coordinate heterogeneous policies for long-horizon tasks without task-specific orchestration co-training?
\textbf{RQ4:} Can RoboHarness solve memory-dependent tasks requiring information beyond the current observation?
\textbf{RQ5:} Can RoboHarness be extended to long-horizon mobile manipulation tasks?
\textbf{RQ6:} How does each design component contribute to overall performance?

\textbf{RQ1: Capability-aware policy routing.}
To evaluate whether RoboHarness can adapt policy routing to context-dependent capability changes, we use $\pi_{0.5}$ as the default base policy and pair it with a TAMP policy that provides complementary structured execution. On LIBERO-Plus, we compare the VLA invocation ratio with its standalone success rate across perturbation categories (\Cref{fig:policy_usage}.(a)). VLA usage generally increases in categories with stronger standalone performance, providing behavioral evidence that routing responds to context-dependent reliability. The VLA is used more frequently under lighting, background, and layout changes than under robot-state perturbations. Sensor noise is an exception as it blurs observations, impairing the understanding skill's assessment.

\textbf{RQ2: Robustness to perturbations.}
RoboHarness achieves the highest average success rate of \(91.9\%\), ranking first in six of seven perturbation categories and outperforming every underlying policy overall (\Cref{tab:libero_plus}). These results show that RoboHarness effectively routes tasks according to the relative strengths of different policies, substantially improving robustness. Sensor noise remains the main bottleneck: degraded observations impair the understanding skill's assessment of policy suitability, leaving RoboHarness at \(90.3\%\).

\textbf{RQ3: Coordination of heterogeneous policies for long-horizon tasks.}
On LIBERO-LoHo, RoboHarness achieves the highest progress score and success rate on every task, averaging \(98.2\%\) and \(96.0\%\), respectively (\Cref{fig:policy_usage}.(b)). Individual policies often make partial progress but rarely complete the full task, while hierarchical guidance alone remains substantially below RoboHarness. This gap shows that long-horizon execution requires more than task decomposition: policy capabilities must be matched to individual subtasks, and intermediate states must remain compatible at policy transitions. RoboHarness addresses both requirements via capability-aware routing and Memory Bridge.

\textbf{RQ4: Memory-dependent task execution.}
RMBench distinguishes between \(M(1)\) tasks requiring a single key historical observation and \(M(n)\) tasks requiring information accumulated across multiple past observations. RoboHarness achieves success rates of \(77.0\%\) and \(77.5\%\), respectively, outperforming Mem-0 at both levels, with a larger advantage on \(M(n)\) tasks (\Cref{tab:rmbench}). Across all nine tasks, it succeeds in 139 of 180 episodes, yielding an overall success rate of \(77.2\%\). Its comparable performance across the two levels suggests that it can retain and use multiple historical observations without the degradation observed in the baselines. Task decomposition confines execution to tractable subproblems, while skill-relevant histories provide observations needed for later decisions. Together, these mechanisms support memory-dependent execution beyond immediate visual feedback. Complete task-level results are provided in \Cref{tab:rmbench_per_task}.

\textbf{RQ5: Long-horizon mobile manipulation.}
We evaluate RoboHarness on six selected BEHAVIOR-1K tasks involving long-horizon mobile manipulation (\Cref{tab:b1k}). RoboHarness achieves a success rate of \(0.433\) and a progress score of \(0.658\), exceeding RLC~\citep{larchenko2025behavior} in success and G0.5~\citep{liu2026g05} in progress. These tasks require coordinating navigation, grasping, placement, and precise interactions such as button pressing across spatially distributed locations. The results support the value of combining complementary capabilities within a single execution framework, while the remaining failures highlight the challenge of reliable coordination over long task horizons. 

\begin{table*}[t]
\centering
\caption{RMBench and BEHAVIOR-1K evaluations. RMBench reports success rates (\%) under different memory complexity levels, \(M(1)\) and \(M(n)\). BEHAVIOR-1K reports the average success rate (SR) and progress score (PS); a dash indicates unavailable SR. The best and second-best results for each metric are shown in \textbf{bold} and \underline{underlined}, respectively.}
\label{tab:rmbench_b1k}
\vspace{-2mm}
\begin{subtable}[t]{0.49\linewidth}

\centering
\caption{RMBench results.}
\label{tab:rmbench}

\scriptsize
\setlength{\tabcolsep}{2pt}
\renewcommand{\arraystretch}{1.15}

\begin{tabular*}{\linewidth}{@{\extracolsep{\fill}}lcc@{}}
\toprule
Method & \(M(1)\) & \(M(n)\) \\
\midrule
Diffusion Policy~\citep{diffusionpolicy}
    & 6.4 & 5.0 \\
\(\pi_{0.5}\)~\citep{black2025pi_}
    & 14.4 & 5.5 \\
X-VLA~\citep{zheng2025x}
    & 11.8 & 7.3 \\
Mem-0~\citep{rmbench}
    & \underline{52.8} & \underline{28.5} \\
\midrule
\textbf{RoboHarness (ours)}
    & \(\mathbf{77.0} \)
    & \(\mathbf{77.5} \) \\
\bottomrule
\end{tabular*}
\end{subtable}
\hfill
\begin{subtable}[t]{0.49\linewidth}
\centering
\caption{BEHAVIOR-1K results.}
\label{tab:b1k}

\scriptsize
\setlength{\tabcolsep}{2pt}
\renewcommand{\arraystretch}{1.15}

\begin{tabular*}{\linewidth}{@{\extracolsep{\fill}}lcc@{}}
\toprule
Method & SR & PS \\
\midrule
\(\pi_{0.5}\)~\citep{black2025pi_}
    & 0.05 & 0.26 \\
RLC~\citep{larchenko2025behavior}
    & \underline{0.23} & 0.42 \\
X-VLA~\citep{zheng2025x}
    & 0.05 & 0.15 \\
G0.5~\citep{liu2026g05}
    & -- & \underline{0.54} \\
\midrule
\textbf{RoboHarness (ours)}
    & \(\mathbf{0.43}\)
    & \(\mathbf{0.66}\) \\
\bottomrule
\end{tabular*}
\end{subtable}
\vspace{-4mm}
\end{table*}

\begin{figure}[t]
\centering
\includegraphics[width=\linewidth]{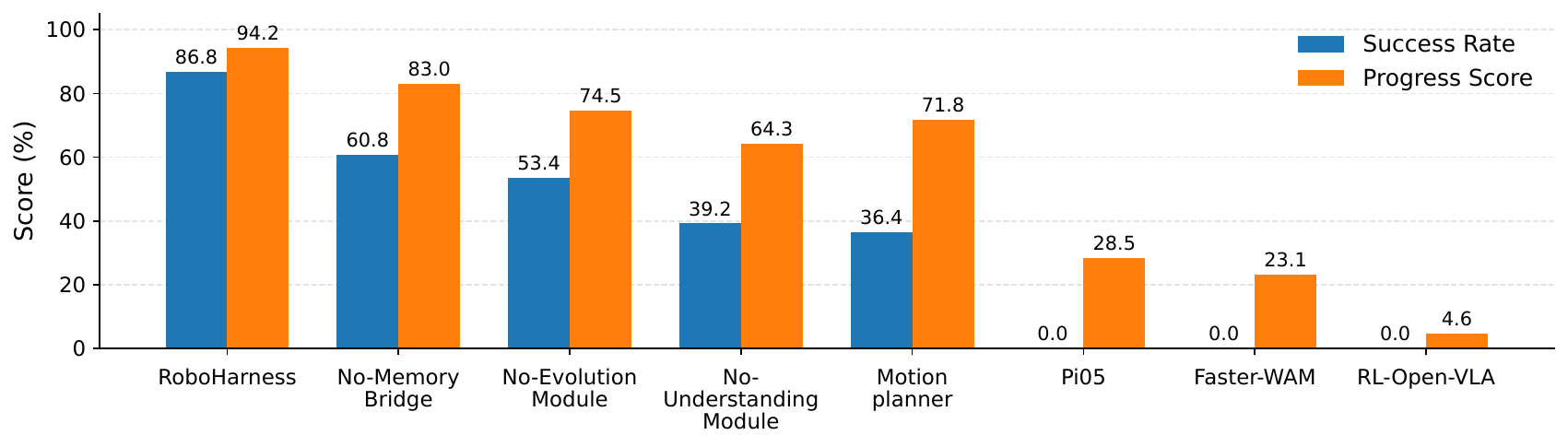}
\caption{Ablation study of RoboHarness on LIBERO-Complex. We compare average success rates and progress scores for the full system and variants with individual components removed.}
\vspace{-4mm}
\label{fig:ablation_study}
\end{figure}


\textbf{RQ6: Ablation Study.} RoboHarness outperforms each constituent policy on 500 randomized instances across 10 long-horizon task classes designed for heterogeneous policy coordination (\Cref{fig:ablation_study}). Among the three component ablations, removing understanding skills causes the largest performance drop, highlighting the value of contextual assessment for task decomposition and policy routing. Removing evolution skills also reduces performance, supporting the contribution of online refinement as experience accumulates. Without Memory Bridge, the system retains substantial partial progress but completes fewer tasks. This pattern is consistent with Memory Bridge helping successful subtasks carry through to full-task completion by improving policy handoffs. Together, these results support the complementary roles of contextual assessment, online adaptation, and memory-guided handoffs.

\section{Real Robot Experiments}

\begin{figure}[t]
\centering
\includegraphics[width=1.0\linewidth]{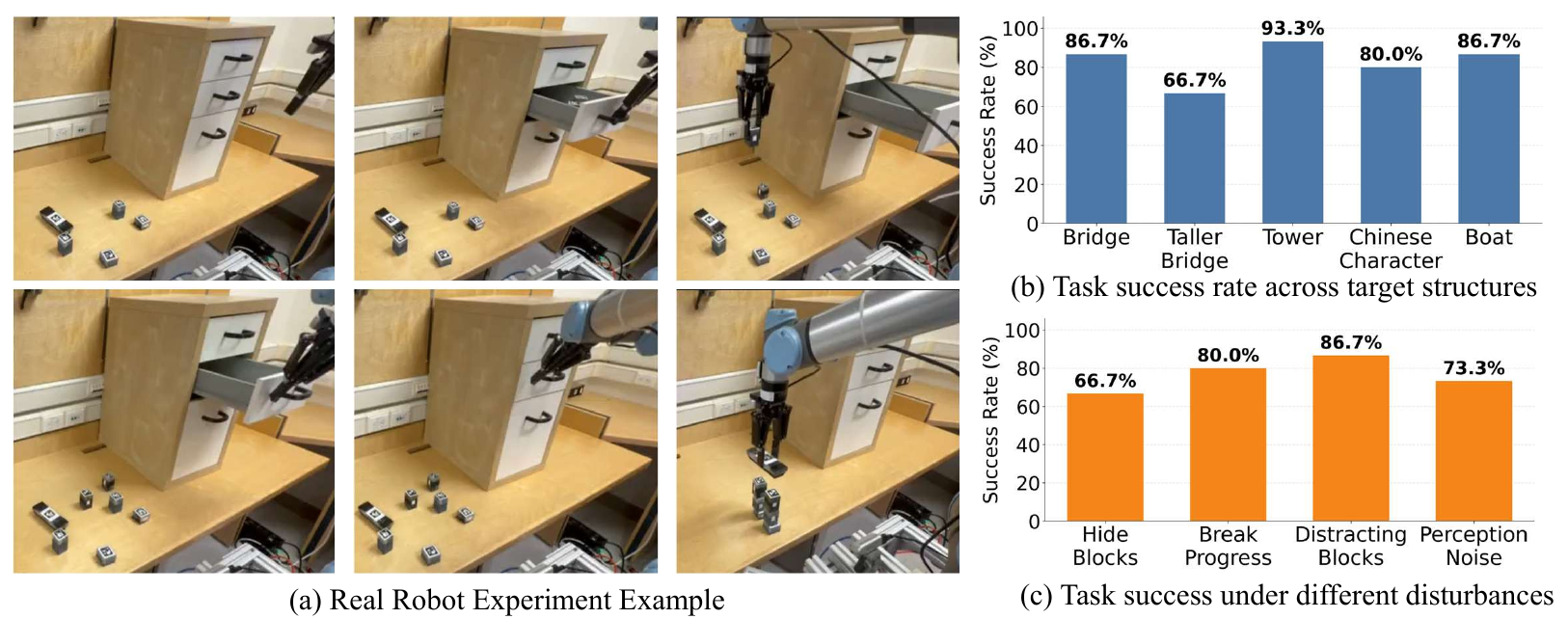}
\caption{(a) Real robot experiment examples. (b) Overall success Rate of various target structure. (c) Overall success rate under different disturbances}
\label{fig:real-robot}
\vspace{-4mm}
\end{figure}

We conducted 135 real-robot trials: 15 for each of five structures and four disturbance settings. In each construction trial, one to three required blocks were randomly hidden in a cabinet, requiring VLA-based identification and retrieval before TAMP assembly. As shown in \Cref{fig:real-robot}, performance was consistent across structures, with lower success on the Taller Bridge due to its complexity and physical instability. RoboHarness tolerated partial dismantling, pose estimation errors, and distractors, while re-hiding blocks caused the largest degradation by requiring repeated exploration and replanning. Timeouts were the primary failure mode. Under our setting, reasoning and planning API calls incurred an average latency of \(4.6\) seconds each, contributing \(67.4\) seconds of additional latency per trial on average across all 135 trials. This cumulative overhead reflects the multiple reasoning calls required for tasks comprising \(20\text{--}30\) subtasks.

\section{Limitations and Future Work}

Although RoboHarness effectively coordinates heterogeneous policies, it remains bounded by their capabilities and cannot perform tasks requiring capabilities absent from the entire policy library. Its reasoning and planning stages also add to overall execution time. Future work will broaden the policy library and use RoboHarness to collect execution trajectories for further training of robot policies. Incorporating additional modalities, such as tactile feedback, may improve contextual assessment.

\section{Conclusion}

We presented RoboHarness, a unified framework that couples capability-aware policy assignment with memory-guided handoffs for long-horizon manipulation. Contextual assessments, policy metadata, and the execution log support decomposition and routing, while Memory Bridge uses the incoming policy's retrieved experience to select reachable, memory-supported handoff configurations. The framework requires no task-specific orchestration pretraining or joint policy retraining, while permitting online adaptation and experience accumulation during evaluation. Results across five public benchmarks, including selected BEHAVIOR-1K tasks, 500 randomized instances across 10 task templates, and 135 real-robot trials demonstrate the effectiveness of the complete system. Component ablations further support the complementary value of contextual assessment, online refinement, and bridging for heterogeneous policy orchestration.

\newpage
\bibliography{iclr2027_conference}
\bibliographystyle{iclr2027_conference}

\newpage
\appendix

\section{LIBERO-Complex Customized Tasks}
\label{task_detail}

In this section, we provide full details of the LIBERO-Complex benchmark used in this paper, including both simulated and real-robot tasks.

\subsection{Simulated tasks}

We construct 500 simulated instances from 10 long-horizon task templates, each evaluated under 50 randomly initialized tabletop configurations. The templates combine ambiguous or state-dependent instructions with object rearrangement, articulated-object manipulation, spatial reasoning, multi-object handling, and state-changing interactions. Approximately 20\% of the ground-truth subtasks can be completed by only one of the four policy types—TAMP, VLA, WAM, or RL—given their respective capabilities. The remaining subtasks have overlapping policy coverage, but the policies differ in their success rates on them. The templates therefore test both whether RoboHarness selects a policy with the required capability and whether it selects an effective policy when multiple options are available. Results are reported in \Cref{fig:ablation_study}. The ten templates are listed below:

\begin{enumerate}
    \item Open the top drawer of the cabinet, place the middle bowl inside it, and place the remaining bowls on the plate.

    \item Place the butter and chocolate pudding in the pan, turn on the stove, and finally place the pan on the stove.

    \item Stack the bowl on the cabinet onto the ramekin, place the bowl on the stove onto the plate, and place the cookies on the cabinet.

    \item Place both bowls on the microwave and close the microwave.

    \item Place the bowl in the top drawer, close the drawer, and place the ketchup on the plate.

    \item Place the ramekin on the stove, retrieve the Akita black bowl from the open top drawer of the wooden cabinet and place it on the plate, and place the cookies in the open drawer.

    \item Place the bowl on the stove, turn on the stove, place the cream cheese on the bowl, and place the wine on the rack.

    \item Collect the cookies, alphabet soup, cream cheese, and butter, and place them in the tray.

    \item Place the side bowl on the cabinet, the center bowl on the plate, and the ramekin on the stove.

    \item Switch the position of the left bowl and popcorn, and collect the other bowl on the cabinet.
\end{enumerate}

These tasks test several challenges beyond standard single-stage manipulation. First, expressions such as ``the middle bowl,'' ``the remaining bowls,'' ``the side bowl,'' and ``the already-open top drawer'' require grounding objects through their spatial relations and current scene state rather than fixed object identities. Second, the mixed and long-horizon requirement require the system to coordinate different underlying heterogeneous policies to accomplish the full task. Third, many instructions contain ordering dependencies: containers must be opened before objects are inserted, appliances must be activated at the appropriate stage, and intermediate placements alter the scene for subsequent subtasks. Finally, the combination of semantic interaction, geometric placement, and multi-step task reasoning requires RoboHarness to alternate between policies with complementary capabilities and maintain execution consistency across their handoffs.

\section{Real robot Experiment}

\begin{figure}[t]
\centering
\includegraphics[width=1.0\linewidth]{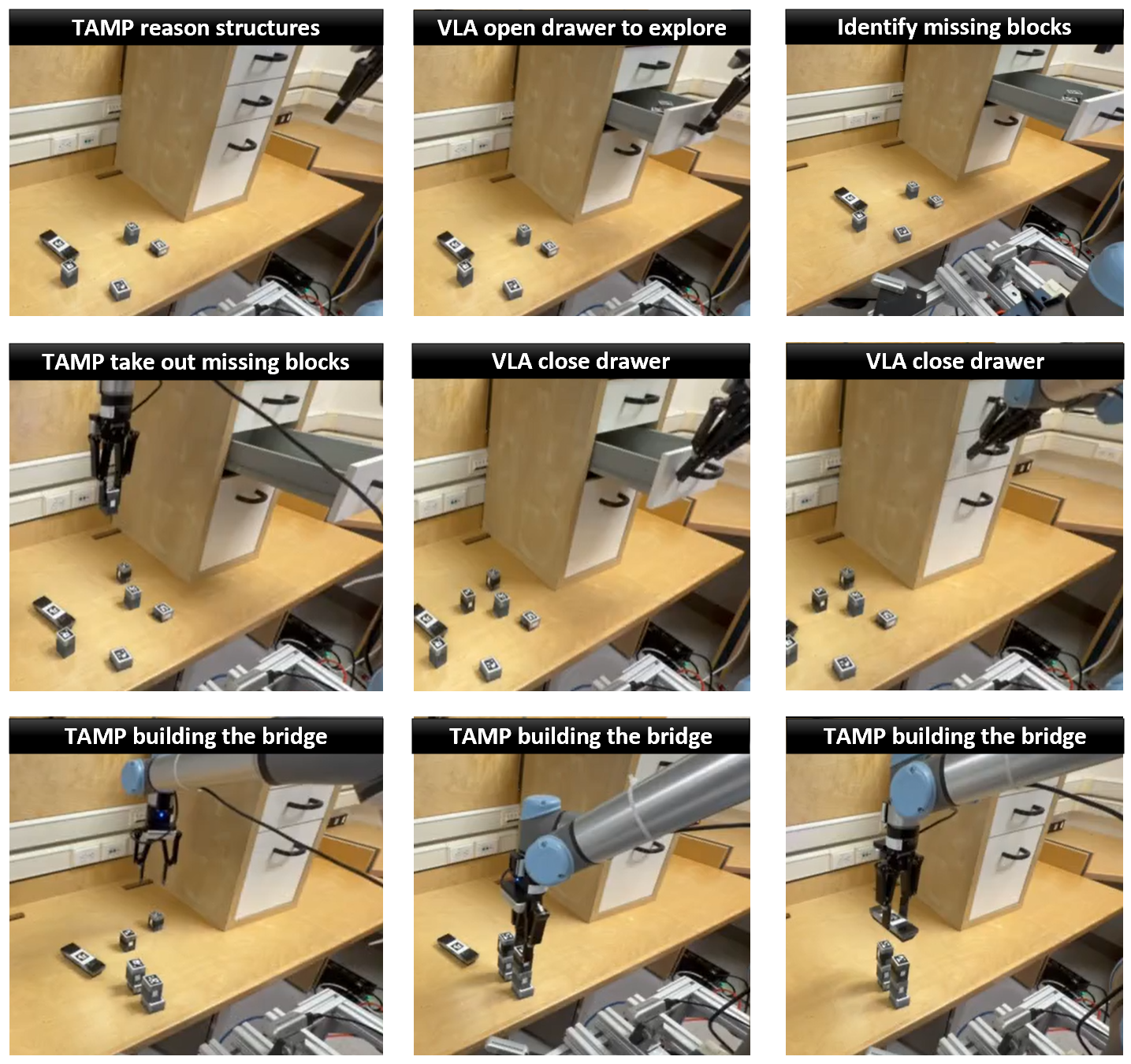}
\caption{Full process of the real-robot experiment}
\label{fig:full process}
\end{figure}

\begin{figure}[t]
\centering
\includegraphics[width=\linewidth]{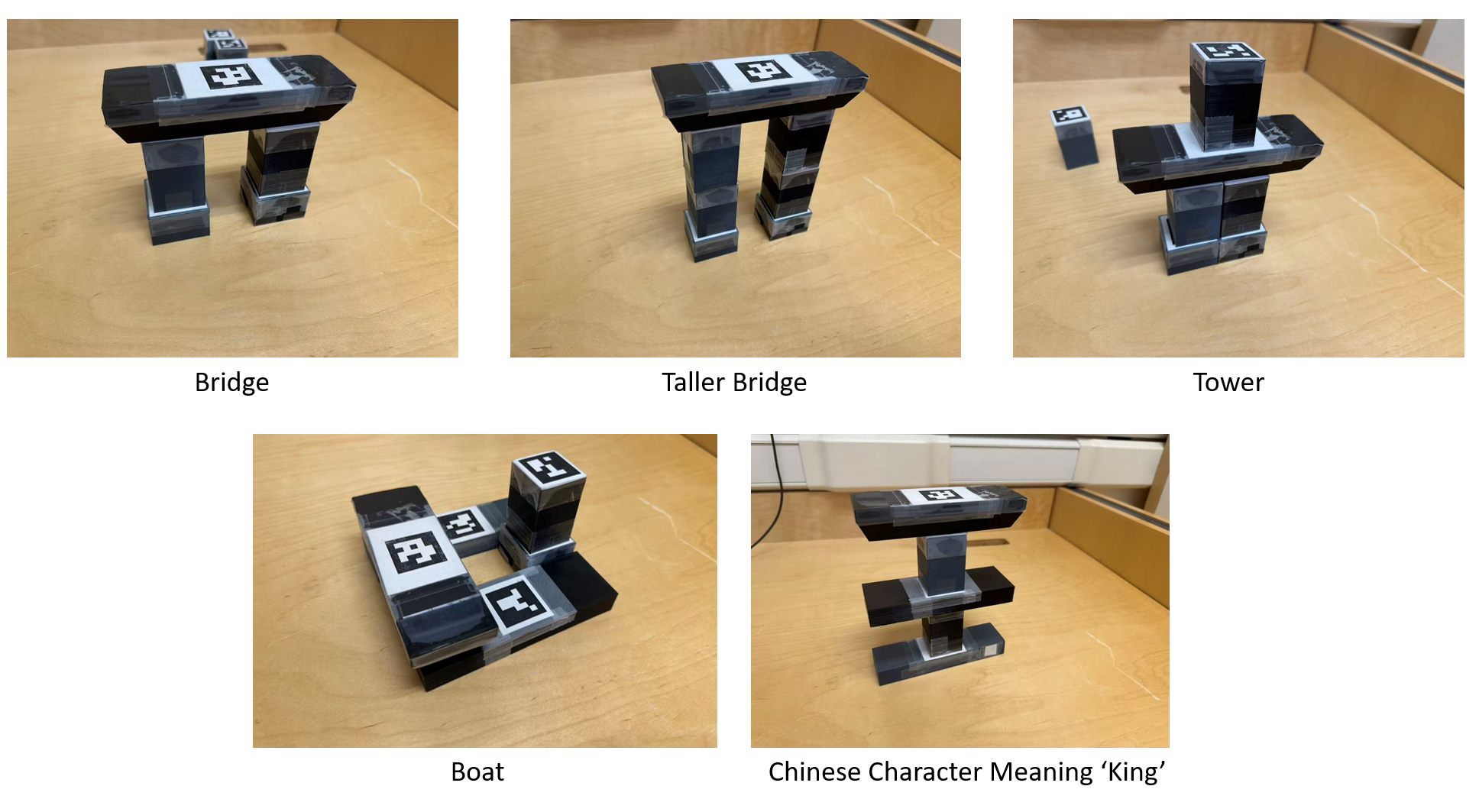}
\caption{Five different target structure}
\label{fig:5_struct}
\end{figure}

\paragraph{Five different target structures:} Using UR5e, We conducted 135 real-robot trials across five structure-construction tasks and four execution-time disturbance settings, with 15 trials for each setting. The full process of the experiment is shown in \Cref{fig:full process}. The five task classes are \emph{Bridge}, \emph{Taller Bridge}, \emph{Tower}, \emph{Chinese Character}, and \emph{Boat} (\Cref{fig:5_struct}). The \emph{Bridge} uses short blocks as lower piers, long blocks as vertical supports, and flat blocks as the bridge deck. The \emph{Taller Bridge} increases the structural complexity by stacking two layers of long blocks to support an elevated deck. The \emph{Tower} is a three-level structure constructed by arranging short and long blocks in a staggered configuration. For the \emph{Chinese Character} task, one of four characters， is randomly selected as the target structure for each trial. The \emph{Boat} is constructed by stacking blocks from short to long to form the hull and cabin, with a long block placed vertically on top as the mast.

In each trial, RoboHarness receives a language instruction describing the target structure and observes the blocks available in the workspace. It must infer the required block configuration, identify missing components, and generate a long-horizon assembly plan. Some required blocks are initially hidden inside a storage cabinet, requiring the robot to open the cabinet, actively explore its contents, and retrieve the missing blocks before completing the structure.

Our real-robot system integrates a TAMP planner with a VLA policy, $\pi_{0.5}$, fine-tuned on 50 human-demonstrated trajectories each for opening and closing cabinet drawers. The TAMP planner performs structural reasoning and generates the geometrically precise sequence of pick-and-place operations required for assembly. When RoboHarness determines that a required block is hidden, it transfers control to $\pi_{0.5}$ through the Memory Bridge. The VLA handles contact-rich interactions that are difficult to model explicitly within TAMP, including opening, closing the cabinet, and exploring its contents. Control is then transferred back to TAMP, which updates the object state and resumes the remaining assembly sequence. This workflow requires repeated capability-aware routing and stable handoffs between semantic, contact-rich exploration and precise long-horizon construction.

\paragraph{Execution-Time Disturbances: }
We further evaluate robustness by introducing four disturbances during execution: (1) re-hiding required blocks after the initially hidden blocks have been retrieved, forcing RoboHarness to detect the newly missing components and repeat the exploration procedure; (2) dismantling a partially completed structure, requiring the system to reassess task progress and replan the assembly; (3) injecting 5\%-10\% relative percentage noise into object-pose estimates to test robustness to perception errors; and (4) introducing distracting objects into the workspace to evaluate task-relevant reasoning and object selection. These disturbances test whether RoboHarness can recognize unexpected state changes, update its plan, reroute subtasks to appropriate policies, and recover without restarting the entire task.

\paragraph{Implementation Details.}
Our real-robot system is implemented on a UR5e manipulator and integrates a TAMP policy with a $\pi_{0.5}$ VLA policy. Two RealSense D435 cameras provide visual observations: one captures the agent view, and the other is mounted on the robot wrist. The TAMP policy uses the UR5e low-level control API for robot motion, PDDLStream~\citep{Garrett2020-cr} for symbolic task reasoning and continuous motion-parameter sampling, and an ArUco-marker-based perception pipeline for object-pose estimation. It reasons about the required blocks and executes geometrically constrained pick-and-place operations during structure construction. The VLA policy is initialized from the official \texttt{pi05\_libero} checkpoint and fine-tuned using 50 human-demonstrated trajectories for each of drawer opening and drawer closing. It handles the contact-rich cabinet interactions required to explore the drawer.

\section{Details of Auxiliary Skill Modules}
\label{implement}

This section provides implementation details corresponding to the understanding skills in \Cref{sec:understanding_skill}, memory management and Memory Bridge in \Cref{sec:memory_skill}, self-evolution skills in \Cref{sec:evolution_skill}, and heterogeneous policy interfaces in \Cref{sec:policy_skill}.

\subsection{Understanding Skills}
\label{app:understanding_skills}

\begin{itemize}
    \item \textbf{Uncertainty Assessment.}
    Perception outputs are aggregated over a temporal window to estimate the mean and variance of object poses. These statistics characterize temporal stability and are provided to the coding agent to support uncertainty-aware planning and policy selection.

    \item \textbf{Visual-Context Assessment.}
    We use two frozen visual encoders from Hugging Face:
    \texttt{google/siglip2-base-patch16-224} and
    \texttt{facebook/dinov2-base}.
    Let \(e_{\mathrm{sig}}\) and \(e_{\mathrm{dino}}\) denote their embedding functions. The visual similarity used in \Cref{sec:memory_management} is
    \[
    \begin{aligned}
    \operatorname{sim}_{\mathrm{vis}}(o,o')
    ={}&
    \alpha\operatorname{cos}\!\left(
    e_{\mathrm{sig}}(o),e_{\mathrm{sig}}(o')
    \right)\\
    &+(1-\alpha)\operatorname{cos}\!\left(
    e_{\mathrm{dino}}(o),e_{\mathrm{dino}}(o')
    \right),
    \end{aligned}
    \]
    where \(\alpha\in[0,1]\) controls the relative contribution of the encoders. This score is used for visual retrieval and provides evidence about the similarity between the current observation and policy-specific execution contexts.

    \item \textbf{Semantic-Context Assessment.}
    We use the frozen Hugging Face language encoder
    \texttt{BAAI/bge-large-en-v1.5} as \(e_{\mathrm{text}}\).
    Task instructions and candidate subtasks are embedded and compared with stored descriptions using cosine similarity. The resulting scores support semantic retrieval and assessment of a policy's task relevance.

    \item \textbf{Input-Quality Assessment.}
    Observation quality is evaluated using conventional image-processing algorithms. Exposure is measured from the mean grayscale intensity and the proportions of pixels below and above predefined intensity thresholds. Sharpness is quantified by the variance of the Laplacian response, while contrast is measured by the standard deviation of grayscale intensities. Image entropy is computed from the intensity histogram to characterize intensity variation. The normalized metrics are combined into an overall quality score and provided to the coding agent for observation selection and reliability-aware decision-making.

    \item \textbf{State-Policy Compatibility Assessment.}
    For a candidate policy and subtask, RoboHarness retrieves relevant policy-specific memory and constructs the local support region and progress estimator described in \Cref{sec:memory_bridge}. The current robot state is assessed using its distance to retrieved states and its local progress score. These outputs provide evidence about robot-configuration compatibility and possible bridging requirements; they are interpreted jointly with visual, semantic, and task-state information by the coding agent.
\end{itemize}

\subsection{Memory Skill Implementation Details}
\label{app:memory_skills}

The memory skill library supports memory retrieval, memory update, and Memory Bridge construction. The retrieval and target-selection mechanisms are defined in \Cref{sec:memory_skill}. Here, we describe the stored data, the progress estimator used in our experiments, and a representative bridge execution.

\begin{figure}[t]
\centering
\includegraphics[width=\linewidth]{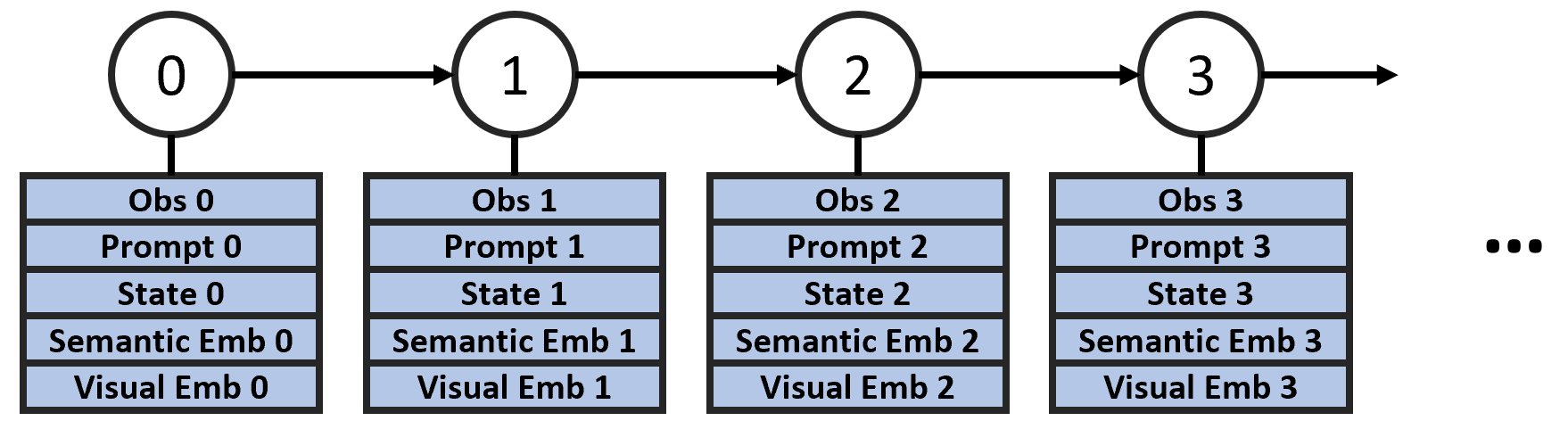}
\caption{Memory Data Structure. Successful policy executions are stored in policy-specific episodic memory banks as temporally linked nodes containing observations, active subtask instructions, robot states, gripper states, and visual and semantic embeddings. Links support retrieval of execution context before and after an anchor node.}
\label{fig:memory_data_structure}
\end{figure}

\paragraph{Memory Data Structure.}
\label{app:memory_data_structure}

As illustrated in \Cref{fig:memory_data_structure}, each linked-node trajectory records a successful policy execution from its initial observation to its terminal state. Nodes are connected in temporal order, allowing the system to recover preceding and subsequent execution context from a retrieved anchor. Each node stores the current visual observation \(o(n)\), active subtask instruction \(g(n)\), raw robot state \(\widetilde{\mathbf{s}}(n)\), gripper state \(z(n)\), and corresponding visual and semantic embeddings.

The raw robot state extracted from the full system state \(\mathbf{x}\) is
\[
\widetilde{\mathbf{s}}(\mathbf{x})
=
\begin{bmatrix}
\mathbf{p}^{\top} &
\mathbf{q}^{\top} &
\boldsymbol{\theta}^{\top}
\end{bmatrix}^{\top},
\]
where \(\mathbf{p}\in\mathbb{R}^{3}\) is the end-effector position in meters, \(\mathbf{q}\in\mathbb{R}^{4}\) is its unit-quaternion orientation, and \(\boldsymbol{\theta}\) contains the joint angles in radians. At bridge-construction time \(t\), we normalize the raw state using the element-wise ranges of the corresponding policy-specific memory bank:
\[
\mathbf{s}_t(\mathbf{x})
=
\frac{\widetilde{\mathbf{s}}(\mathbf{x})-\boldsymbol{\ell}_t}
{\max\!\left(\mathbf{u}_t-\boldsymbol{\ell}_t,\epsilon\right)},
\]
where \(\boldsymbol{\ell}_t\) and \(\mathbf{u}_t\) are the element-wise minimum and maximum of the raw states stored in that bank at time \(t\), \(\epsilon>0\) prevents division by zero, and the division and maximum are applied element-wise. We use these same ranges to normalize the current state and all retrieved memory states for a given bridge construction. We write \(\mathbf{s}(n)\) as shorthand for \(\mathbf{s}_t(\mathbf{x}(n))\) when \(t\) is fixed. A current state outside the memory-bank range may have normalized components outside \([0,1]\). Euclidean distances are computed between states normalized using the same ranges.

Visual embeddings from the frozen SigLIP2 and DINOv2 encoders and semantic embeddings from the frozen language encoder are stored alongside the node data to support efficient retrieval.

The execution log \(\mathcal{H}\) maintains a complete chronological trace of the orchestration process. Each entry records the task and scene context, agent or skill invocation, exchanged messages and outputs, observations, actions, runtime feedback and errors, bridge executions, and intermediate or final outcomes. It also records evolution actions and their results, including parameter changes, metadata revisions, interface or routing updates, validation outcomes, and whether each change is retained. Evolution skills retrieve relevant log segments as evidence for capability assessment, failure diagnosis, and subsequent adaptation. If a policy execution succeeds, its recorded nodes are additionally linked chronologically and inserted as an episodic trajectory into the corresponding policy-specific bank \(\mathcal{M}_i\). Contextual trajectory retrieval and Memory Bridge construction operate only on successful episodic trajectories in these banks.

\paragraph{Progress Estimator.}
\label{app:progress_estimator}

We implement \(f_{\mathrm{score},t}\) using a linear pairwise ranking support vector machine (SVM). For each bridge construction, the retrieved anchor nodes are expanded into local trajectory segments as described in \Cref{sec:memory_bridge}, preserving the same subtask and gripper state. Each sample consists of a normalized robot-state feature vector \(\mathbf{s}_{i,j}=\mathbf{s}(n_{i,j})\) and a signed temporal offset \(y_{i,j}=\Delta t_{i,j}\). Before training, the temporal offsets are normalized to \([-1,1]\):
\[
\hat{y}_{i,j}
=
\frac{y_{i,j}}{\Delta t_t^{\max}},
\qquad
\Delta t_t^{\max}
=
\max_{i,j}\left|y_{i,j}\right|,
\qquad
\mathcal{D}_t
=
\left\{
\left(\mathbf{s}_{i,j},\hat{y}_{i,j}\right)
\right\}.
\]
The anchor has \(y_{i,0}=0\), subsequent nodes have positive offsets, and preceding nodes have negative offsets.

The normalized temporal labels define relative preferences rather than regression targets. Because the anchors are retrieved using visual and semantic similarity, we treat them as approximately aligned points of task progress and use their relative temporal offsets to construct preferences across the retrieved segments. For two samples \(a\) and \(b\) in \(\mathcal{D}_t\), \(\hat{y}_a>\hat{y}_b\) indicates that \(\mathbf{s}_a\) should receive a higher progress score than \(\mathbf{s}_b\). We construct the pair set
\[
\mathcal{P}_t
=
\left\{
(a,b)\mid \hat{y}_a>\hat{y}_b
\right\}
\]
and fit the ranking SVM by solving
\[
\begin{aligned}
\min_{\mathbf{w}_t,\,\{\xi_{ab}\}}
\quad&
\frac{1}{2}\|\mathbf{w}_t\|_2^2
+
C_{\mathrm{svm}}
\sum_{(a,b)\in\mathcal{P}_t}\xi_{ab},
\\
\text{subject to}\quad&
\mathbf{w}_t^{\top}
\left(\mathbf{s}_a-\mathbf{s}_b\right)
\geq
1-\xi_{ab},
\\
&
\xi_{ab}\geq0,
\qquad
\forall(a,b)\in\mathcal{P}_t,
\end{aligned}
\]
where \(C_{\mathrm{svm}}\) controls the trade-off between regularization and ranking violations. The resulting progress estimator is
\[
f_{\mathrm{score},t}(\mathbf{s})
=
\mathbf{w}_t^{\top}\mathbf{s}.
\]

Thus, the output represents an ordering of local progress rather than elapsed time, a probability density, or a probability of successful execution. Although the temporal labels are normalized to \([-1,1]\) before training, they impose only relative ordering constraints; consequently, the learned SVM score is not itself restricted to this range. In particular, the zero label assigned to an anchor does not force its learned score to be zero.

Candidate targets are evaluated using the learned score together with the memory-support constraint, motion-planning feasibility, and motion cost defined in \Cref{sec:memory_bridge}. We initialize the acceptance threshold as \(\theta_{\mathrm{score},t}=0.9\). If candidate sampling produces no memory-supported, motion-feasible state whose score exceeds the current threshold, we reduce \(\theta_{\mathrm{score},t}\) by \(0.02\) and repeat the sampling and filtering procedure until an acceptable candidate is found or the 10-second time limit is reached. Fitting this auxiliary ranking model uses retrieved experience and does not jointly retrain the underlying control policies.

\begin{figure}[t]
\centering
\includegraphics[width=\linewidth]{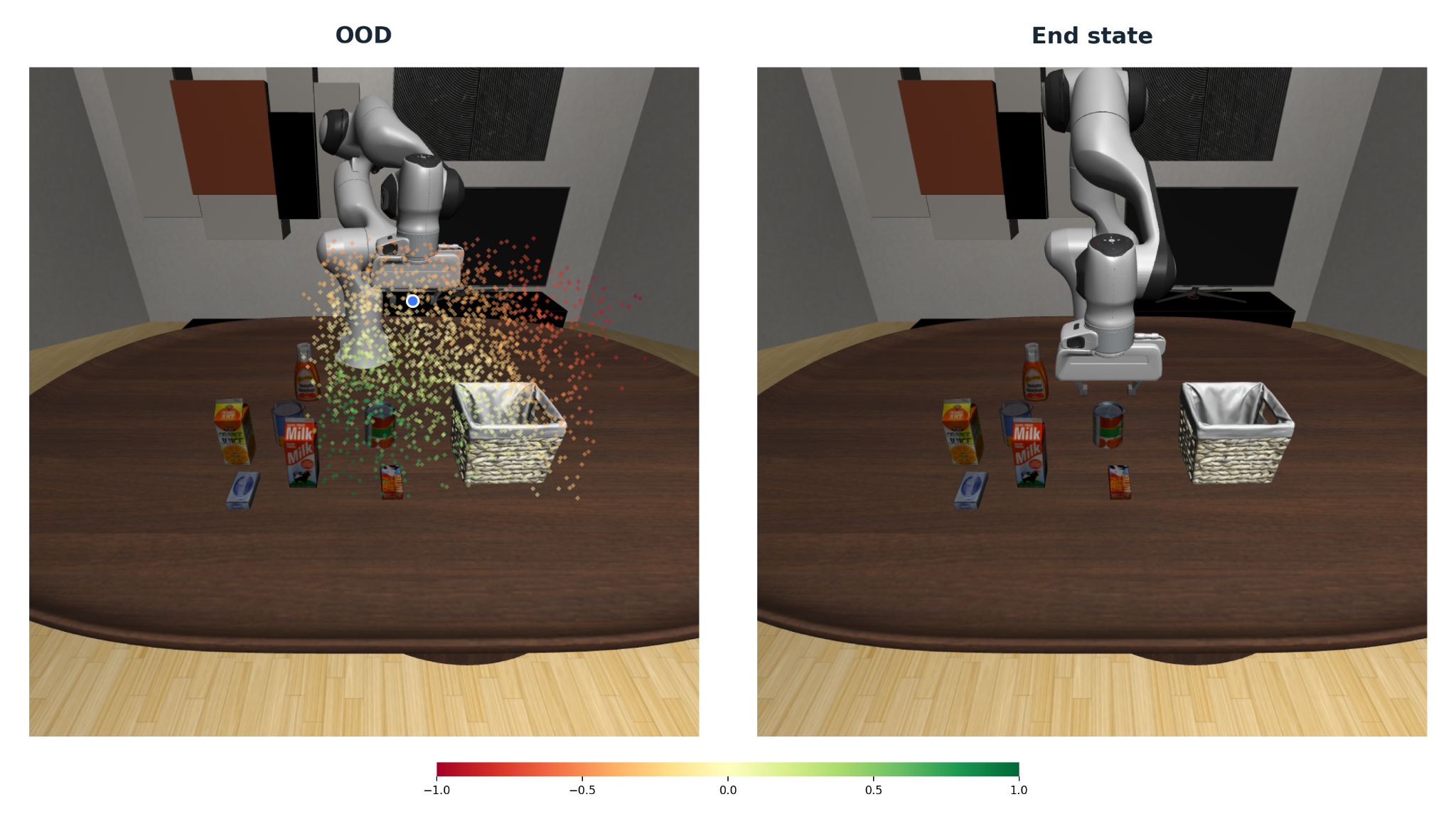}
\caption{A Memory Bridge handoff. Left: an out-of-distribution (OOD) robot configuration and sampled candidate states, colored by their relative state scores (red: lower; green: higher). Right: the robot configuration goes back to distribution after executing the memory bridge trajectory.}
\label{fig:memory_bridge_example}
\end{figure}

\paragraph{Memory Bridge Example.}
\label{app:memory_bridge_example}

\Cref{fig:memory_bridge_example} shows a representative policy handoff. The preceding policy leaves the robot in a configuration poorly aligned with the incoming policy's execution experience. 
Candidate states are filtered by memory support, progress score, and motion feasibility. RoboHarness selects a handoff target by balancing progress against motion cost, with \(\lambda_{\mathrm{motion}}=0.2\), and plans a bridge to that target. After execution, the coding agent reassesses the observation and task context before invoking the next policy or revising the routing decision. This handoff requires no joint retraining of the constituent policies.

\subsection{Evolution Skill Implementation Details}
\label{app:evolution_skills}

The self-evolution skill library contains three components: policy adaptation, harness refinement, and metadata update. These components use execution outcomes and retrieved experience to improve individual policies and their orchestration.

\begin{itemize}
    \item \textbf{Policy Adaptation.}
    We implement SIMPACT~\citep{liu2026simpact} and PDDLLM~\citep{Huang2025-ue} as adaptation methods for the TAMP system. SIMPACT adjusts motion-level parameters, such as object grasp poses, using execution feedback. PDDLLM extends the symbolic task model by learning new logical representations from observed interactions. These methods support adaptation to previously unsupported objects, relations, and task conditions. Policy adaptation also includes tuning exposed policy hyperparameters using the parameter-update procedure described below.

    \item \textbf{Harness Refinement.}
    RoboHarness uses Codex with GPT-5.5 to refine orchestration code online. Based on execution traces, runtime errors, and discrepancies between expected and observed outcomes, the coding agent identifies implementation bugs, repairs policy interfaces, and revises policy-routing and coordination logic. Modified implementations are evaluated through subsequent execution before being retained. Harness refinement also adjusts orchestration parameters, including retrieval counts, Memory Bridge support and score thresholds, and bridge-trajectory acceptance criteria.

    \item \textbf{Metadata Update.}
    Policy capability metadata is revised using execution evidence in \(\mathcal{H}\). It records contexts in which a policy performs reliably or fails, including relevant objects, scenes, task types, and environmental conditions. Updated metadata is reflected in the policy cards and recorded in \(\mathcal{H}\) for subsequent planning, routing, and evolution.
\end{itemize}

Parameter updates within Policy Adaptation and Harness Refinement share a directional tuning procedure. Each tunable parameter is assigned a discrete grid of candidate values. The coding agent selects whether to increase, decrease, or retain the current setting, and a separate tuning module maps this decision to the corresponding grid value and applies the update. This implements directional search over discretized parameters rather than exhaustive grid search.

All three components are enabled in the simulated experiments. Persistent failures trigger the retrieval of relevant execution evidence from \(\mathcal{H}\), after which the coding agent selects the appropriate updates. The selected adaptations, their validation outcomes, and retained changes are appended to the log. Validated changes are retained across evaluation episodes, allowing continual adaptation. Thus, the setting does not require task-specific pretraining of the orchestration framework but permits the use of experience accumulated during evaluation.

\subsubsection{Underlying Heterogeneous Policies}
\label{app:underlying_policies}

We incorporate four independently developed policies with complementary capabilities. Each is exposed through a callable skill interface and a policy card, as described in \Cref{sec:policy_skill}, while retaining its native inputs, outputs, and execution procedure.

\begin{itemize}
    \item \textbf{\(\boldsymbol{\pi_{0.5}}\).}
    For LIBERO, LIBERO-Plus, and LIBERO-LoHo, we use the official \texttt{pi05\_libero} checkpoint released by Physical Intelligence through the \texttt{openpi} repository. This checkpoint was fine-tuned on LIBERO-Spatial, LIBERO-Goal, LIBERO-Long, and LIBERO-Object, providing broad language-conditioned manipulation capabilities~\citep{black2025pi_}. For RMBench and BEHAVIOR-1K, we use checkpoints trained by us for the respective benchmarks. For our \(\pi_{0.5}\) fine-tuning, we adapt the official \texttt{openpi} training configuration\footnote{\url{https://github.com/Physical-Intelligence/openpi/blob/main/src/openpi/training/config.py}} to the task-specific data. We use an action horizon of 50, a global batch size of 256, and the learning rate warms up for 10,000 steps to \(5\times10^{-5}\) and then decays over the next 20,000 steps. We use the checkpoint at step 30,000.

    \item \textbf{Faster-WAM.}
    We incorporate Faster-WAM~\citep{ma2026faster} as a world-action-model policy and train it on LIBERO using the authors' released training pipeline. Its Dock of Transformer architecture connects a lightweight action head to a video Transformer backbone, allowing action prediction to use representations from the video model. RoboHarness invokes Faster-WAM through its native inference interface, adding video-model-based control to the heterogeneous policy library without requiring a shared action representation or joint retraining with the other policies.

    \item \textbf{RL-Post-Trained OpenVLA-OFT.}
    We use the RLinf/RLinf-OpenVLAOFT-GRPO-LIBERO-90 checkpoint released on Hugging Face. This policy was post-trained with GRPO on the 90 short-horizon tasks in LIBERO-90 using the RLinf framework~\cite{zang2025rlinf}. It represents the RL-post-trained policy family within RoboHarness.

    \item \textbf{TAMP Planner.}
    We implement a TAMP system for geometrically constrained manipulation over a predefined object set. PDDL represents object types, predicates, and manipulation actions with their preconditions and effects, while each problem instance specifies the available objects, initial state, and task goal. We use the Fast-Forward (FF) planner~\citep{Hoffmann2001-fy} through PDDLStream~\citep{Garrett2020-cr}, which connects symbolic planning with Python sampling procedures for continuous quantities such as grasp poses, placement poses, and robot configurations.

    High-level actions, including picking, placing, stacking, and aligning, are executed using three motion generation and control mechanisms: (1) pose-based motion planning, which estimates object poses and generates robot trajectories using RRT through OMPL~\cite{guo2026open}; (2) VLM-based visual control, which guides robot motion from visual observations by outputting grid-based adjustments in eight translational and six rotational directions; and (3) memory-based trajectory augmentation, which incorporates retrieved demonstration segments into execution. In the memory-based configuration, memory skills retrieve successful episodic trajectories from \(\mathcal{M}\) to disambiguate visually identical states, and each policy invocation is preceded by replaying the leading fraction (\(\beta=0.4\)) of a retrieved demonstration segment. These mechanisms provide alternative ways to ground symbolic actions, while the TAMP system remains constrained by its predefined task domain and available execution capabilities.
\end{itemize}

Together, these policies provide complementary language-conditioned, video-model-based, reinforcement-learned, and symbolic-geometric capabilities for evaluating heterogeneous policy orchestration. Policy cards record their interface requirements, supported tasks, assumptions, constraints, and execution statistics, and are updated as new evidence accumulates.

\section{Evaluation Details on Benchmarks}
\label{evals}

\subsection{Heterogeneous Policy Configurations across Benchmarks}
\label{app:policy_benchmark}

\Cref{tab:policy_configurations} summarizes the heterogeneous policies available to RoboHarness across benchmarks and ablation studies. Policy configurations are determined by the availability of compatible checkpoints and the requirements of each benchmark. For LIBERO, LIBERO-Plus, RMBench, and BEHAVIOR-1K, RoboHarness orchestrates $\pi_{0.5}$ and TAMP, combining learned visuomotor control with explicit task and motion planning. For LIBERO-LoHo and the ablation study, the policy library additionally includes Faster-WAM and RLinf-OpenVLA, enabling evaluation with all four policy families. A checkmark denotes that a policy is available to the orchestrator, rather than necessarily invoked in every task; the policies used during execution depend on capability-aware task decomposition and routing.

\begin{table}[h]
\centering
\caption{Heterogeneous policy configurations used by RoboHarness across benchmarks and ablation studies. A checkmark indicates that the policy is available for orchestration.}
\label{tab:policy_configurations}
\resizebox{\linewidth}{!}{%
\begin{tabular}{@{}lcccc@{}}
\toprule
Benchmark / Study & $\pi_{0.5}$ & Faster-WAM & RLinf-OpenVLA & TAMP \\
\midrule
LIBERO & $\checkmark$ & -- & -- & $\checkmark$ \\
LIBERO-Plus & $\checkmark$ & -- & -- & $\checkmark$ \\
LIBERO-LoHo & $\checkmark$ & $\checkmark$ & $\checkmark$ & $\checkmark$ \\
RMBench & $\checkmark$ & -- & -- & $\checkmark$ \\
BEHAVIOR-1K & $\checkmark$ & -- & -- & $\checkmark$ \\
Ablation Study & $\checkmark$ & $\checkmark$ & $\checkmark$ & $\checkmark$ \\
\bottomrule
\end{tabular}%
}
\end{table}

\subsection{RMBench}
\label{rmben}

RMBench~\citep{rmbench} is a memory-dependent robotic manipulation benchmark built on RoboTwin 2.0, comprising nine dual-arm manipulation tasks across two levels of task memory complexity: five $M(1)$ tasks and four $M(n)$ tasks. The $M(1)$ tasks require retaining information from a single key observation or a fixed, limited set of historical frames, such as an object's original position. The $M(n)$ tasks require accumulating and reasoning over multiple past observations, such as tracking previous attempts and their outcomes during trial-and-error interactions. These tasks evaluate memory retention and utilization when the current observation alone is insufficient for decision-making. We evaluate RoboHarness on 20 episodes per task, with each episode recorded as a binary success or failure.

\begin{table*}[h]
\centering
\caption{Per-task success rates (\%) on RMBench. Diffusion Policy, $\pi_{0.5}$, X-VLA, and Mem-0 results are from the RMBench paper~\citep{rmbench}; RoboHarness results use our full nine-task evaluation.}
\label{tab:rmbench_per_task}
\small
\setlength{\tabcolsep}{5pt}
\renewcommand{\arraystretch}{1.08}
\resizebox{\textwidth}{!}{%
\begin{tabular}{@{}llccccc@{}}
\toprule
Task & TMC & Diffusion Policy & $\pi_{0.5}$ & X-VLA & Mem-0 & \textbf{RoboHarness} \\
\midrule
Observe and Pick Up & $M(1)$ & 1 & 9 & 9 & 4 & 90 \\
Rearrange Blocks & $M(1)$ & 0 & 13 & 13 & 89 & 85 \\
Put Back Block & $M(1)$ & 0 & 11 & 18 & 90 & 100 \\
Swap Blocks & $M(1)$ & 11 & 24 & 16 & 67 & 95 \\
Swap T & $M(1)$ & 20 & 15 & 3 & 14 & 15 \\
\midrule
Battery Try & $M(n)$ & 10 & 16 & 26 & 28 & 50 \\
Blocks Ranking Try & $M(n)$ & 10 & 6 & 1 & 18 & 80 \\
Cover Blocks & $M(n)$ & 0 & 0 & 2 & 68 & 80 \\
Press Button & $M(n)$ & 0 & 0 & 0 & 0 & 100 \\
\midrule
\textbf{Overall average} & -- & 5.8 & 10.4 & 9.8 & 42.0 & \textbf{77.2} \\
\bottomrule
\end{tabular}%
}
\end{table*}

\subsection{BEHAVIOR-1K}
\label{behavior1k}

BEHAVIOR-1K~\citep{li2024behavior1k} is a large-scale benchmark for evaluating embodied agents on diverse, realistic household activities in interactive environments. We select six mobile-manipulation tasks with varying horizons and interaction requirements: Turning On Radio (Task 0), Picking Up Trash (Task 1), Setting Mousetraps (Task 5), Moving Boxes to Storage (Task 16), Tidying Bedroom (Task 18), and Putting Shoes on Rack (Task 22). Our evaluations of RoboHarness, $\pi_{0.5}$~\citep{black2025pi_}, Robot Learning Collective (RLC)~\citep{larchenko2025behavior}, and X-VLA~\citep{zheng2025x} use 10 episodes per task. For G0.5~\citep{liu2026g05}, we use the published per-task progress scores of its four-epoch model; per-task binary success rates are unavailable. These tasks require agents to navigate across environments, manipulate multiple objects, and execute temporally ordered subtasks under substantial viewpoint and background changes. We report the arithmetic mean across the six tasks and evaluate SR independently from PS using the benchmark's binary success criterion. The task-level results are provided in \Cref{tab:behavior_per_task}.

\begin{table}[h]
\centering
\caption{Per-task results on the six selected BEHAVIOR-1K tasks. SR and PS denote success rate and progress score, respectively. Dashes mark unavailable results. The average is the arithmetic mean across tasks, and all results are rounded to two decimal places.}
\label{tab:behavior_per_task}
\small
\setlength{\tabcolsep}{4pt}
\renewcommand{\arraystretch}{1.08}
\resizebox{\textwidth}{!}{%
\begin{tabular}{@{}lcccccccccc@{}}
\toprule
& \multicolumn{2}{c}{\textbf{RoboHarness}} & \multicolumn{2}{c}{$\pi_{0.5}$} & \multicolumn{2}{c}{RLC} & \multicolumn{2}{c}{X-VLA} & \multicolumn{2}{c}{G0.5} \\
\cmidrule(lr){2-3}\cmidrule(lr){4-5}\cmidrule(lr){6-7}\cmidrule(lr){8-9}\cmidrule(l){10-11}
Task & SR & PS & SR & PS & SR & PS & SR & PS & SR & PS \\
\midrule
Task 0: Turning On Radio & 0.90 & 0.90 & 0.20 & 0.20 & 0.30 & 0.30 & 0.20 & 0.20 & -- & 0.05 \\
Task 1: Picking Up Trash & 0.50 & 0.73 & 0.00 & 0.15 & 0.30 & 0.62 & 0.10 & 0.40 & -- & 0.85 \\
Task 5: Setting Mousetraps & 0.40 & 0.67 & 0.00 & 0.23 & 0.10 & 0.36 & 0.00 & 0.12 & -- & 0.57 \\
Task 16: Moving Boxes to Storage & 0.60 & 0.75 & 0.10 & 0.33 & 0.70 & 0.73 & 0.00 & 0.03 & -- & 0.55 \\
Task 18: Tidying Bedroom & 0.00 & 0.43 & 0.00 & 0.33 & 0.00 & 0.20 & 0.00 & 0.08 & -- & 0.62 \\
Task 22: Putting Shoes on Rack & 0.20 & 0.46 & 0.00 & 0.34 & 0.00 & 0.33 & 0.00 & 0.05 & -- & 0.57 \\
\midrule
\textbf{Average} & \textbf{0.43} & \textbf{0.66} & 0.05 & 0.26 & \underline{0.23} & 0.42 & 0.05 & 0.15 & -- & \underline{0.54} \\
\bottomrule
\end{tabular}%
}
\end{table}

\newpage

\subsection{LIBERO-LoHo}
\label{liberoloho}

LIBERO-LoHo evaluates long-horizon task execution across five tasks using progress score and success rate. RoboHarness achieves an average progress score of \(98.2\%\) and success rate of \(96.0\%\), outperforming all compared methods on both metrics across all five tasks (\Cref{tab:libero_loho}).

\begin{table}[h]
\centering
\caption{LIBERO-LoHo long-horizon evaluation. Each entry reports the progress score / success rate (\%). The best and second-best results for each metric are shown in \textbf{bold} and \underline{underlined}, respectively.}
\label{tab:libero_loho}

\small
\setlength{\tabcolsep}{3pt}
\renewcommand{\arraystretch}{1.05}

\begin{tabular*}{\linewidth}{
    @{\extracolsep{\fill}}lcccccc@{}
}
\toprule
Method & Task 1 & Task 2 & Task 3 & Task 4 & Task 5 & Overall \\
\midrule

H-WM-\(\pi_{0.5}\)
& \underline{98.0 / 94.0}
& \underline{86.7 / 60.0}
& \underline{74.0 / 46.0}
& \underline{70.7 / 42.0}
& \underline{95.0 / 82.0}
& \underline{84.9 / 64.8} \\

Logic-guided \(\pi_{0.5}\)
& 95.3 / 86.0
& 84.7 / 58.0
& 54.7 / 16.0
& 39.3 / 4.0
& 92.0 / 78.0
& 73.2 / 48.4 \\

LLM-guided \(\pi_{0.5}\)
& 84.7 / 54.0
& 80.7 / 42.0
& 68.0 / 24.0
& 41.3 / 4.0
& 59.5 / 10.0
& 66.8 / 26.8 \\

\(\pi_{0.5}\)
& 66.0 / 4.0
& 73.3 / 24.0
& 54.7 / 4.0
& 44.7 / 0.0
& 38.0 / 0.0
& 55.3 / 6.4 \\

\(\pi_0\)
& 54.0 / 0.0
& 62.0 / 28.0
& 44.0 / 0.0
& 31.3 / 0.0
& 34.0 / 0.0
& 45.1 / 5.6 \\

X-VLA
& 48.8 / 0.0
& 16.7 / 0.0
& 23.3 / 0.0
& 33.3 / 0.0
& 40.0 / 0.0
& 32.4 / 0.0 \\

Faster-WAM
& 0.0 / 0.0
& 64.0 / 2.0
& 28.7 / 0.0
& 32.0 / 0.0
& 0.0 / 0.0
& 24.9 / 0.4 \\

OpenVLA-OFT
& 0.0 / 0.0
& 60.0 / 0.0
& 17.3 / 0.0
& 22.7 / 0.0
& 0.0 / 0.0
& 20.0 / 0.0 \\

GR00T-N1.5
& 0.0 / 0.0
& 30.7 / 0.0
& 34.7 / 0.0
& 2.7 / 0.0
& 0.0 / 0.0
& 13.6 / 0.0 \\

\midrule
\textbf{RoboHarness (ours)}
& \textbf{99.3 / 98.0}
& \textbf{97.3 / 96.0}
& \textbf{99.3 / 98.0}
& \textbf{98.0 / 96.0}
& \textbf{97.0 / 92.0}
& \textbf{98.2 / 96.0} \\

\bottomrule
\end{tabular*}
\end{table}

\end{document}